\documentclass{article}
\usepackage{spconf,amsmath,graphicx}
\usepackage[inkscapelatex=false]{svg}
\usepackage{threeparttable}
\usepackage{multirow}
\usepackage{amsmath,amsfonts}
\usepackage[colorlinks, linkcolor=blue,citecolor=blue]{hyperref}
\usepackage{marvosym}
\usepackage{enumitem}
\setlist{nosep, leftmargin=14pt}

\usepackage{mwe} 

\title{Brighteye: Glaucoma Screening with Color Fundus
Photographs based on Vision Transformer}
\name{Hui Lin  \textsuperscript{\Letter}\thanks{\Letter \
 huilin2023@u.northwestern.edu}, Charilaos Apostolidis, Aggelos K. Katsaggelos}
\address{Department of Electrical and Computer Engineering, Northwestern University, Evanston, IL, USA}
\begin{document}
%
\maketitle
\begin{abstract}
Differences in image quality, lighting conditions, and patient demographics pose challenges to automated glaucoma detection from color fundus photography. Brighteye, a method based on Vision Transformer, is proposed for glaucoma detection and glaucomatous feature classification. Brighteye learns long-range relationships among pixels within large fundus images using a self-attention mechanism. Prior to being input into Brighteye, the optic disc is localized using YOLOv8, and the region of interest (ROI) around the disc center is cropped to ensure alignment with clinical practice. Optic disc detection improves the sensitivity at 95\% specificity from 79.20\% to 85.70\% for glaucoma detection, and the Hamming distance from 0.2470 to 0.1250 for glaucomatous feature classification. In the developmental stage of the Justified Referral in AI Glaucoma Screening (JustRAIGS) challenge, the overall outcome secured the fifth position out of 226 entries. 
\end{abstract}
\begin{keywords}
Glaucoma screening, color fundus photography, optic disc detection, vision transformer, feature aggregation, image classification
\end{keywords}
\section{Introduction}
\label{Introduction}

Fundus photographs play a crucial role in the early detection and monitoring of glaucoma due to their ability to visualize the optic nerve head and retinal structures. However, manual glaucoma detection demands considerable labor and necessitates specialized clinical expertise. Therefore, there is a strong interest in developing highly accurate and robust detection methods suitable for clinical applications.

Convolutional neural networks (CNNs) have presented a promising performance in multiple applications in manufacturing \cite{lin2019automated,wang2022using} as well as healthcare \cite{lin2023stenunet,liu2023yolo}.
In medical image applications, transformers have been applied to a wide variety of tasks, such as left atrium segmentation \cite{lin2024usformer}. Inspired by the aforementioned methods, we introduce Brighteye\footnote{The code is available at \href{https://github.com/HuiLin0220/brighteye.git}{https://github.com/HuiLin0220/brighteye.git}}, a ViT-based model adapted for glaucoma detection and glaucomatous feature classification from color fundus photography. Before entering Brighteye, YOLOv8 is employed to locate the optic disc, and subsequently, the region of interest (ROI) surrounding the disc center is cropped. Brighteye is validated in the Justified Referral in AI Glaucoma Screening (JustRAIGS) challenge \cite{lemij2023characteristics} dataset. The comprehensive result was ranked fifth in the development phase among 226 submissions. The Brighteye model’s performance exceeds the minimum sensitivity criteria recommended by Prevent Blindness America \cite{hemelings2023generalizable}.

The remainder of the paper is organized as follows. The details of the proposed network are given in Section \ref{methods}, including information on the architecture and loss function. Datasets, metrics, and implementation details are shown in Section \ref{Experiments}. Experimental results are shown in Section \ref{Results}. Section \ref{Conclusions} presents the conclusions and future directions.

\section{Methods}
\label{methods}
\begin{figure*}[h]
    \centering
    \includegraphics[width=\textwidth]{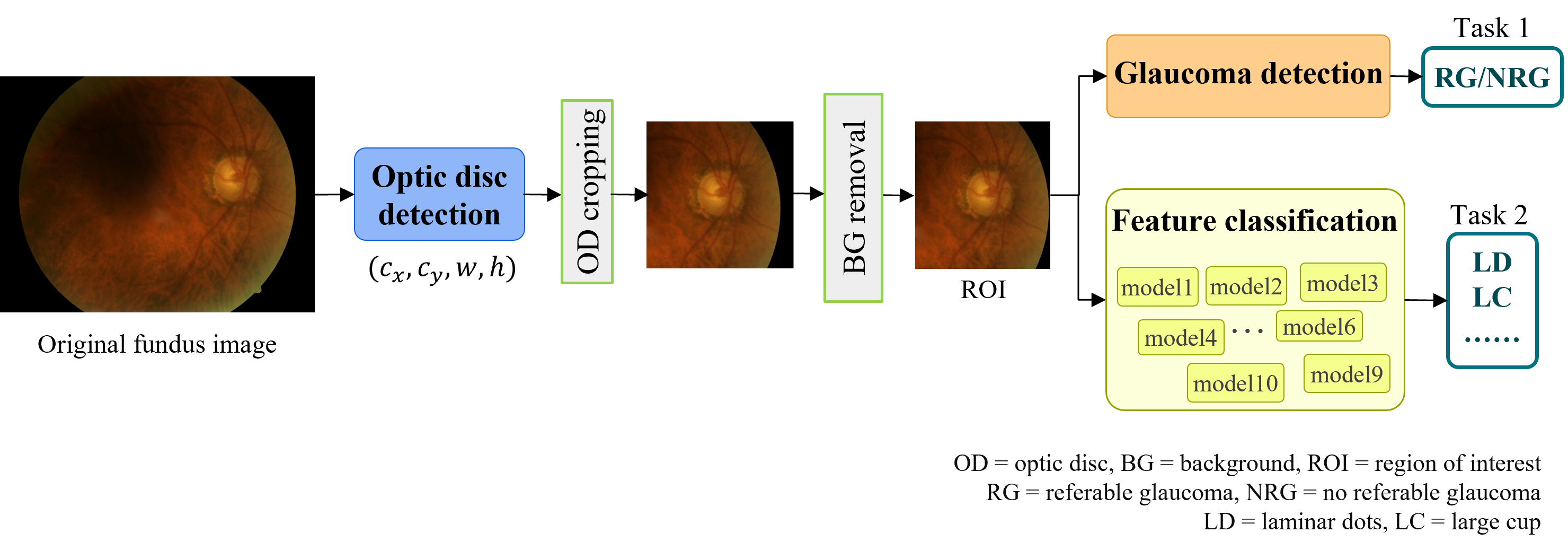}
    \caption{The proposed framework comprising optic disc detection, glaucoma detection, and feature classification. In the first step, a region of interest (ROI) around the detected optic disc (OD) is cropped. In the second step, the presence of glaucoma is predicted for Task 1's output, and 10 binary classifiers predict the presence of 10 individual features independently for Task 2's output.}
    \label{fig: pipeline}
\end{figure*}

Our workflow, illustrated in Fig. \ref{fig: pipeline}, consists of optic disc detection, glaucoma detection, and glaucomatous feature classification. As detailed in Section \ref{detection}, the center coordinates and size of the optic disc (OD) are predicted in the first step. Then, the optic disc region is cropped, and the background is removed. Glaucoma detection is characterized as a binary classifier, whereas glaucomatous feature classification employs 10 independent binary classifiers for independent prediction of the presence of each feature. Altogether, there are 11 binary independent classifiers sharing the same model architecture, as elaborated in Section \ref{Classifier}.

\begin{figure*}[h]
    \centering
    
    \includegraphics[width=0.75\textwidth]{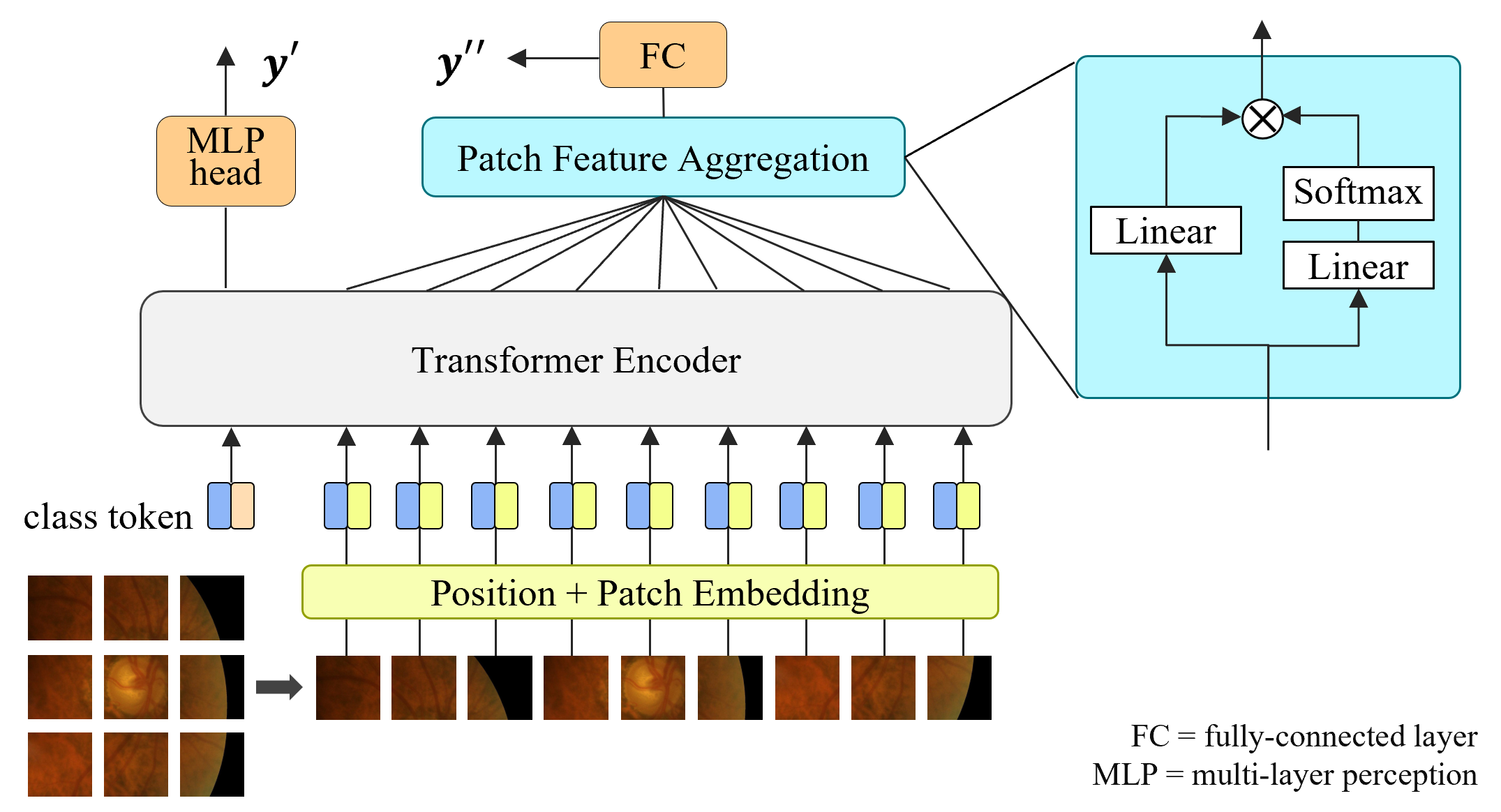}
    \caption{The architecture of Brighteye, modified from ViT \cite{dosovitskiy2021image} for binary classification. All 11 classifiers in Fig. \ref{fig: pipeline} share this architecture.} 
    \label{fig: cls}
\end{figure*}

\subsection{Optic Disc Detection}
\label{detection}
Clinically, ophthalmologists diagnose glaucoma by assessing various features of the optic disc (OD) area, including the neuroretinal rim and the optic cup \cite{septiarini2015automatic,madadi2024domain}.
Inspired by the top-performing methods in the AIROGS challenge \cite{wang2022workflow,khader2022elevating}, optic disc detection is necessary before being fed into downstream tasks. Our further experiments also demonstrate that optic disc detection and cropping significantly enhance the accuracy of glaucoma detection and glaucomatous feature classification, as shown in Table \ref{table: comparison_evaluation}.

YOLOv8 \cite{yolov8_ultralytics} is applied for optic disc detection due to its efficient and accurate object detection capabilities. The predicted center coordinates, width, and height of OD are denoted as $c_x$, $c_{y}$, $w$, and $h$, respectively. Subsequently, a region of interest (ROI) is cropped around the OD center with a size of $(w+h)/2*3$, as long as the optic disc is detected. Otherwise, the original image is fed downstream.

\begin{table*}[h]
\begin{center}
\caption{Evaluation of Brighteye with different combinations of preprocessing for glaucoma detection and glaucomatous feature classification.}
\begin{threeparttable}
\label{table: comparison_evaluation}
\end{threeparttable}
\begin{tabular}{ |c|c|c|c|c|c|c|}
\hline
\multirow[c]{2}{*}{ Experiments } & \multicolumn{2}{|c|}{Preprocessing}&  \multicolumn{2}{|c|}{ TPR@95(\%)} & \multicolumn{2}{|c|}{NHD}\\
\cline{2-7}
&OD cropping& BG removal& Self-test&Development &Self-test &Development\\
\hline
1& &&89.91&79.20&0.1849&0.2470\\
\hline
2&\checkmark&&90.21&83.10&0.1643&0.1510\\
\hline
3&\checkmark&\checkmark&\textbf{91.59} &\textbf{85.70}&\textbf{0.1417}&\textbf{0.1250}\\
\hline
\end{tabular}
\begin{tablenotes}
        \footnotesize
        \item $TPR@95$: sensitivity at 95\% specificity; $NHD$: normalized hamming distance
        \item OD: optical disc, BG: background, $Bold$: best results
\end{tablenotes}
\end{center}
\end{table*}

\begin{figure*}[h]
    \centering
    
    \includegraphics[width=0.9\textwidth]{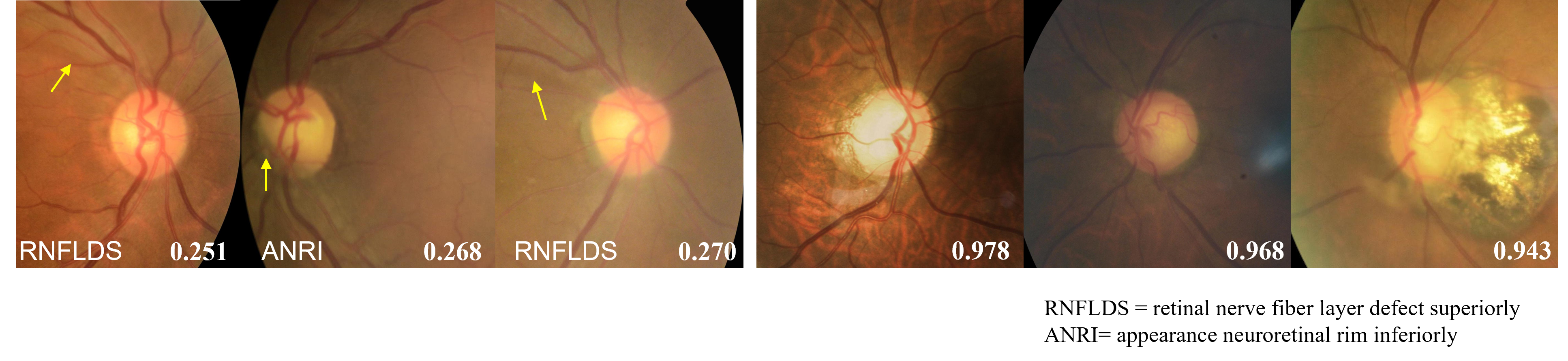}
    \caption{Top three most extreme false-negative (first three columns) and false-positive (last three columns) cases. The value displayed in the bottom right corner indicates the probability of glaucoma. The bottom left corner in the first three columns shows the glaucomatous features identified by the graders, with the area indicated by the yellow arrows.
}
    \label{fig: FP&FN}
\end{figure*}

\subsection{Classification based on Vision Transformer}
\label{Classifier}
All 11 binary classifiers in the proposed framework share the same architecture modified from ViT \cite{dosovitskiy2021image}, as pictured in Fig.\ref{fig: cls}.
The input image with a size of $H\times W$ is divided into several patches of size $P\times P$. The number of patches from a single fundus image is $N = H\times W/P^2$, where $P$ is set as 16 in the following experiments. Following the basics of ViT, each patch is flattened and linearly projected into a 1D vector with $D$ dimensions. This is referred to as patch embedding. To preserve the spatial information, position embedding is added to the patch embeddings, which is encoded as a 1D trainable vector. Moreover, a learnable class token is added to the beginning of the embeddings from patches to serve as a global representation of the whole image. The class token at the output of the transformer encoder is fed to the multi-layer perceptron (MLP) head for binary classification.


In our work, a patch feature aggregation is added to the end of the transformer encoder to aggregate all the features extracted from patches. The weight for each patch is learnable and obtained from two linear projections and the softmax function. To prevent overfitting, layer normalization and ReLU activation are used after each linear projection. The softmax function ensures that all weights sum up to 1. Finally, the aggregated feature is fed to a fully-connected layer for binary classification.



In summary, Brighteye produces two classification results ($y^{\prime},y^{\prime \prime}$) from the class token and aggregated patch features. The training loss $\mathcal{L}$ is the average of two binary cross-entropy losses, as shown in Equation \ref{equation:loss}, and the prediction is averaged during testing.

\begin{equation}
\label{equation:loss}
    \mathcal{L}(y, y^{\prime}, y^{\prime \prime}) = -\sum_{i=1}^2 y_i \log (y_i^{\prime})-\sum_{i=1}^2 y_i \log(y_i^{\prime \prime})
\end{equation}
where $y$ is the ground truth. $y^{\prime}$ and $y^{\prime \prime}$ are the model outputs. $y\in \{0, 1\}$. 1 represents the presence of glaucoma and the corresponding feature in glaucoma detection and feature classification, respectively.

\section{Experiments}
\label{Experiments}

Since the optic disc annotation is not available in the provided dataset, we manually annotated 100 images randomly selected from the database. YOLOv8 was trained on 80 annotations randomly selected and tested in the remaining 20 images. Yolov8's performance is roughly evaluated based on the area under the ROC curve (AUC).


All the ROIs are resized into $512 \times 512$ as model inputs, and the patch size in Brighteye is 16.
The Brighteye model was trained using the Adam optimizer. The learning rate is initially set to 0.0002, and it is adjusted every 5 epochs through multiplication by 0.5.

Given the substantial class imbalance in the challenge dataset, the dataset applied in this paper consists of all referable glaucomatous (RG) cases and 4000 randomly selected non-referable glaucomatous (NRG) cases from the original challenge set. These cases are partitioned into training and validation sets at a 4:1 ratio, with 2616 RG images in the training set and 654 in the validation set.

Data augmentation for classification model training is applied in this work to enhance generalizability and prevent overfitting. A 50\% probability was used to apply horizontal and vertical flips. The rotation angle is randomly selected within the intervals $(-10^{\circ}, 10^{\circ})$. The factors for saturation, brightness, and hue are randomly selected within the intervals $(0.95, 1.05)$. 
\section{Results and Discussion}
\label{Results}


The AUC achieved by the trained optic detection model is 0.995. Even though YOLOv8 is trained on only 80 images, the performance of optic disc detection is promising. Only 0.7\% (670 among 101423) of all challenge images is not detected. The results in Table \ref{table: comparison_evaluation} further demonstrate the role of OD cropping and background removal in enhancing the performance of downstream tasks, resulting in a sensitivity (TPR@95) increase from 79.20\% to 85.70\%, and a Hamming distance decrease from 0.2470 to 0.1250 in the dataset provided in the development phase. The Brighteye model's performance exceeds the minimum sensitivity criteria recommended by Prevent Blindness America.

Figure \ref{fig: FP&FN} shows the top three most extreme false-negative and false-positive cases. The regions pointed by the arrows in the first three columns (RNFLD  and ANRI features) are too shallow to be recognized by classifiers. The fundus images in the last three columns appear noisy, also contributing to the classifier's errors.



\section{Conclusions}
\label{Conclusions}
The study introduces a two-step framework, Brighteye. YOLOv8 locates the optic disc before inputting fundus images into Brighteye, after which the region of interest (ROI) around the disc center is cropped, and the background region is removed. Brighteye is adopted from ViT for its capability to learn the global context in large fundus images through its self-attention mechanism. The framework's performance is ranked 5th in the development stage of the Justified Referral in AI Glaucoma Screening (JustRAIGS) challenge. Our algorithm demonstrates promising performance with a sensitivity of 85.70\% at 95\% specificity for glaucoma detection, and a Hamming distance of 0.1250 for glaucomatous feature classification.




\section{Compliance with Ethical Standards}
This research study was conducted retrospectively using anonymized human subject data made available by the Justified Referral in AI Glaucoma Screening (JustRAIGS) challenge organizers. Ethical approval was not required, as confirmed by the license attached with the open-access data.
\end{document}